# Making informed decisions in cutting tool maintenance in milling: a KNN-based model agnostic approach

**Revati M Wahul[a], Aditya Rahalkar[b], Om Khare[c], Rohan N Soman[d,*], Abhishek Patange[e]**

[a] Department of Computer Engineering, Modern Education Society's Wadia College of Engineering, India
[b] Department of Mechanical Engineering, COEP Technological University, India
[c] Department of Computer and Information Technology, COEP Technological University, India
[d] Institute of Fluid Flow Machinery, Polish Academy of Sciences, Poland
[e] Department of Mechanical Engineering, Cummins College of Engineering for Women, India

## Highlights

- KNN-based white box model for improved interpretability.
- Tuning of hyperparameters using GridsearchCV.
- Reduction in type-2 errors through data augmentation.
- Comparison of contribution of forces from X and Y direction towards tool condition monitoring.

## Abstract

In machining processes, monitoring the condition of cutting tools is crucial to ensure high productivity and consistent product quality. Tool Condition Monitoring (TCM) systems increasingly rely on machine learning techniques to analyze the large volume of sensor data generated during machining operations. In this study, real-time cutting force signals were acquired from milling experiments conducted under multiple tool wear conditions. Statistical feature extraction was performed on the force signals, followed by feature selection using decision tree–based importance analysis. A K-Nearest Neighbors (KNN) classifier was employed for tool condition classification, and hyperparameter tuning was carried out to enhance model performance. A comparative analysis of force signals revealed that the augmented force data in the X-direction achieved superior classification performance, with a maximum accuracy of 96%, compared to 78% for the Y-direction force signals. Data augmentation significantly reduced Type II error, which decreased from 3.04% to 0.14% for the X-direction force data. Hyperparameter tuning further improved model generalization, resulting in a testing accuracy of 95% and a training accuracy of 98% for the tuned KNN model. To enhance transparency and trustworthiness, a model-agnostic white-box interpretability framework was integrated with the KNN classifier, providing both global feature influence analysis and local, instance-level explanations of classification decisions. The proposed approach enables clear identification of dominant force features influencing tool wear classification and supports informed decision-making for tool maintenance. The results demonstrate that combining KNN classification with a model-agnostic interpretability layer offers an effective and transparent solution for force-based Tool Condition Monitoring in milling operations.

## Keywords

tool condition monitoring, k-nearest neighbors, face milling cutter, model agnostic approach, white box model.

(*) Corresponding author.
E-mail addresses: R.M. Wahul (ORCID:0000-0003-2747-7557) rmwahul@mescoepune.org, A. Rahalkar (ORCID:0009-0007-3784-6041) adityarahalkar07@gmail.com, O. Khare (ORCID: 0009-0007-7958-5986) omkhare02@gmail.com, R.N. Soman (ORCID:0000-0002-5499-2565) rsoman@imp.gda.pl, A. Patange (ORCID:0000-0002-9130-694X) abhipatange93@gmail.com

## 1. Introduction

Tool condition monitoring systems play a vital role in modern manufacturing processes, which involve the analysis of the health and performance of different tools, like drills, milling cutters, and grinding wheels, used in machining processes. It is essential to maintain the excellent condition of these tools to ensure product quality and improve tool life. The extreme forces and high temperatures experienced by the tools during machining processes due to improper machining inputs such as speed, feed, and depth of cut, the tools undergo wear and potential damage over time. Recent research has shown that changes in cutting forces, energy consumption, and thermal behavior provide valuable indicators of tool wear and machining stability under varying cutting conditions [1,2]. Since these process responses evolve progressively as tools degrade, monitoring such signals enables early detection of abnormal cutting behavior before severe damage occurs [3,4]. This has motivated the growing use of force-based and data-driven monitoring approaches to better understand tool condition and support reliable machining operations. While designing TCMs, two types of measurement techniques are used viz. direct and indirect measurement. Direct measurement or vision sensor technique involves the use of a camera or an optical system to visualize the wear on the tool. Indirect measurement involves the measurement of different parameters like cutting forces, acoustic emissions, vibrations etc. using numerous sensors. The variation in these parameters in relation to time is used to analyze the condition of the tool. One of the many advantages of TCM is that it reduces unplanned downtime by detecting tool wear in real time. This allows for the well-timed maintenance of the tool and prevents interventions in production schedules which can be costly. TCM is also important from the safety point of view as the malfunctioning tools which can pose safety risks for the workers, can be identified promptly with the help of TCM. Zhou et al. [5] highlighted the importance of tool condition monitoring systems and how they are important for the development of fully automated milling operations. The difficulties related to sensors, feature extraction, monitoring the models and the challenges of the complexity of milling operations are discussed in the paper. Mohanraj et al. [2] examined numerous monitoring methods, such as acoustic emission, cutting force, and vibration signals highlighting the importance of online tool condition monitoring systems and how it is important in cost reduction and quality improvement. Lei et al. [6] explored the evolution of intelligent fault diagnosis in machine fault diagnosis. The study focused on the transition from the traditional machine learning methods to the modern deep learning techniques. It asserted particularly on the KNN algorithm and how it helps in automating machine fault diagnosis and reducing human intervention. Leonhardt et al. [7] presented various classification techniques, from statistical and geometric classifiers to neural networks, highlighting the evolution of automatic diagnosis, and the significance of fuzzy logic in real world diagnostic challenges. TCM is vital for enhancing productivity and cost-effectiveness in machining operations. A concrete and dependable TCM system can lead to significant benefits such as – an increase in cutting speeds by 10 – 50%, a reduction in downtime by scheduling maintenance in advance, and an overall hike in savings ranging between 10 – 40% etc. Such monitoring ensures that tools are operating in optimal conditions, preventing unexpected breakdowns and ensuring consistent product quality [8]. Since the large and complex amount of data generated by different signals can become incomprehensible for humans, machine learning can help a lot in TCM. These techniques can process and analyze the data that may help in predicting the condition of the tool. Serin et al. [9] emphasized TCM as crucial for predicting and avoiding adverse conditions in machining processes. The authors highlight that the inadmissible conditions during machining, such as chatter, tool wear, and deterioration, can directly impact tool life, surface quality, and dimensional accuracy. They also highlight the potential of deep learning methods, such as deep multi-layer perceptron (DMLP) and convolutional neural networks (CNN) in enhancing the prediction and learning capabilities for tool condition monitoring. Table 1 illustrates the past findings of different approaches of applying machine learning for tool condition monitoring.

The gaps identified from the above literature are summarized as follows.

- In most of the approaches, it was observed that a significant portion of the studies leaned towards model-specific approaches in machine learning, rather than adopting a model-agnostic approach. While this approach may yield optimized performance for individual models, it often overlooks the potential benefits of model-agnostic methods, such as improved interpretability and versatility [21].

- A white box model approach increases the transparency in the decision-making process and helps understand how the model makes predictions or decisions, and based on that, judges the model's output. This approach is not used extensively by many. In TCM, having a white box model provides clear insights into the interrelation among the input variables and the output condition of the tool, which is vital in understanding why a tool might be in that condition.
- Hyperparameter tuning is one of the essential factors which determine the model's performance. It finds the best combination of different hyperparameters for which maximum accuracy, minimum error and excellent generalization to the unseen data, is obtained. Tuning the model, will significantly increase the model's performance [22].
- Failing to detect an actual problem in the tool condition (Type 2 error), can be very costly. If the condition of the tool is detected to be positive, when it is actually negative, then it becomes difficult to identify this defect [23]. This can lead to a decrease in the quality of the process and also production efficiency. The importance of detecting type 2 error and a way to reduce it isn't spoken about much.
- The use of real time data plays a pivotal role in TCM. One of the main advantages of using real time data is that it allows the TCMs to detect damage to the tool as it happens and allows the operators to take immediate action. This not only prevents further damage to the tool but also minimizes the downtime by timely scheduling of maintenance of the tool.

Table 1. Different approaches to apply ML in TCM.

| Reference | Objective | Algorithm | Classes | Input signals | Tuning | Explainability |
|---|---|---|---|---|---|---|
| Saglam et al. [10] | Multi NN for TCM in milling based on cutting forces. | Multi Neural Network | 2 (healthy & flank wear) | 6 (velocity, force - Fx, Fy, Fz) | ✓ | × |
| Shankar et al. [11] | Neural Nets for predicting the cutting tools wear | Neural Net & ANFIS | 3 (Fresh, working, dull) | 2 (Resultant force, sound pressure) | ✓ | × |
| Kaya et al. [12] | Use of SVM along with sensor fusion for TCM of milling operation | SVM with sensor fusion | 4 (Sharp, workable, close to dull, dull) | 3 (Cutting Speed, feed per tooth, depth of cut) | × | ✓ |
| Wang et al. [13] | Use of multi-scale PCA for TCM | Multi scale PCA | 2 (Normal, abnormal) | 2 (Force, vibration) | × | × |
| Benkedjouh et al. [14] | Support vector regression for tool health monitoring | SVM | - | 3 (vibrations, force, acoustic emissions) | ✓ | × |
| Torabi et al. [15] | Clustering methods for online TCM for high-speed milling | Clustering method | 3 (Normal, half-worn, worn) | 2 (Servo motor current) | × | × |
| Rao et al. [16] | Self-organizing map for tool wear monitoring | Kohonen's self-organizing map | 2 (In-condition, worn-out) | 1 (radial cutting force) | × | × |
| Dahe et al. [17] | Random forest and FURIA for TCM using statistical analysis | Random Forest & FURIA | 4 (Good, flank wear, thermal cracks, broken) | 3 (Spindle Speed, Tool feed, depth of cut) | × | × |
| Kothuru et al. [18] | CNN for TCM with the help of deep visualization for end milling operations | CNN | 4 (Good, average, advanced, failure) | 4 (spindle speed, chip load, axial depth of cut, radial depth of cut) | ✓ | × |
| Fatemeh et al [19] | CNN and spectral subtraction for TCM in milling operations | CNN, SVM, KNN | - | 3 (Force, Vibration, spindle motor current) | × | × |
| Balachandar et al. [20] | Random forest for TCM of friction stir welding using vibration signals | Decision Tree, Hoeffeding, Random Forest | 3 (Good, air gap, broken) | 1 (Vibration) | × | × |

## 2. Contributions & methodology

Fig. 1 represents a complete flow of the work incorporated in this research.

- In the following paper, a comprehensive approach is used to train on the raw data. Preprocessing techniques like outlier removal from the experiment data, raw data augmentation and feature selection is done to refine the data and to enhance the model's performance.
- Type 2 error as mentioned can lead to serious problems. Efforts are taken to reduce such false positives as much as possible. This was mainly achieved by data augmentation which led to better training, thus enabling the model to better classify the tool condition.

- To make the classification robust, hyperparameter tuning was done to achieve maximum performance from the proposed model.
- A white box model approach was adopted to enable transparency in the model used since the real-life nature of the classification. White-box approach enables us to understand the internal workings and identify the parameters that affect the decision-making process of the classification model.

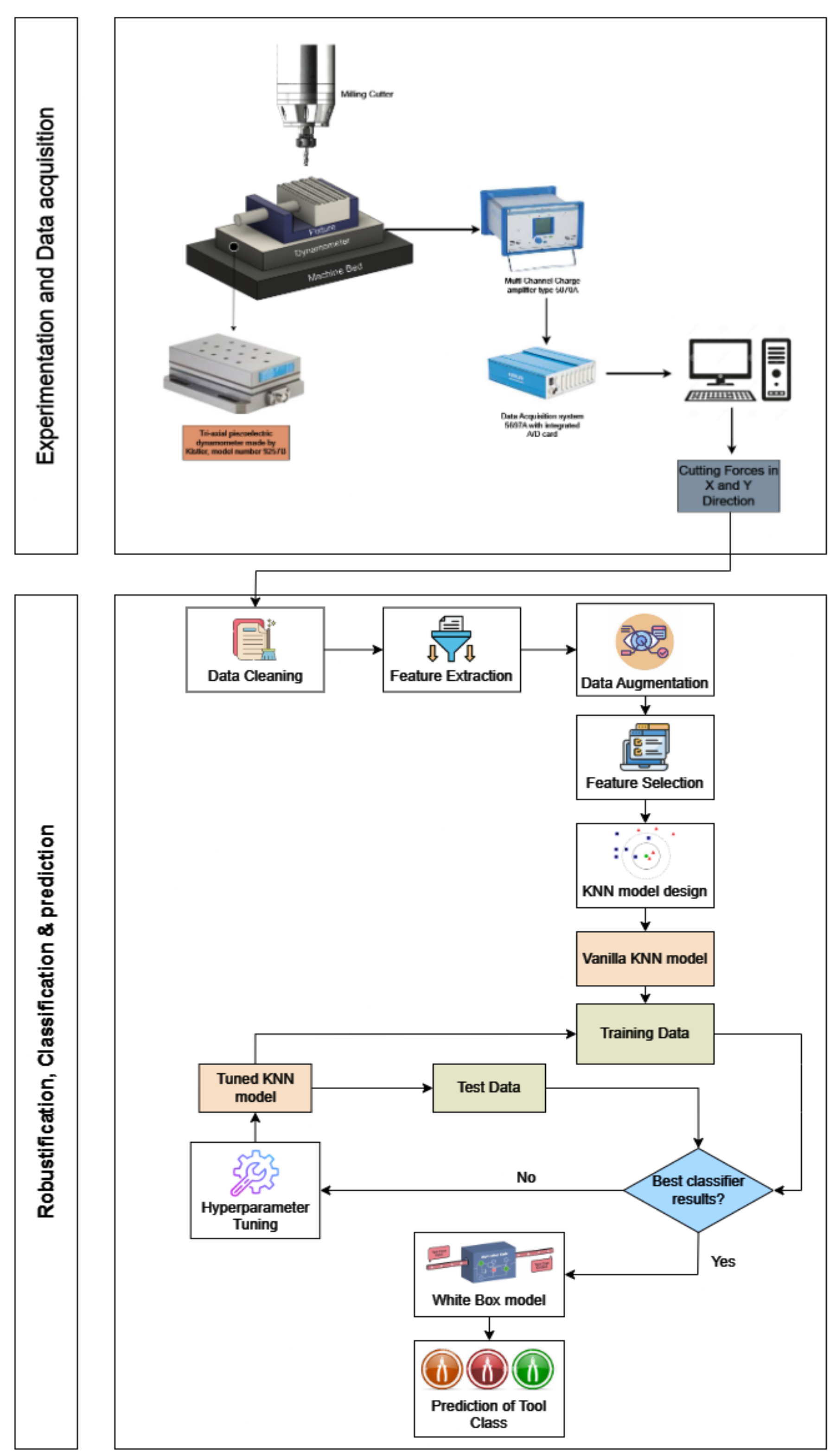

Figure 1. Schematic of data collection and processing.

## 3. Experimentation and data collection

Numerous experiments were performed on a computer numerical control (CNC) vertical machining center equipped with a maximum spindle power of 15 kW and a maximum spindle speed of 15,000 revolutions per minute (rpm). The experiments were designed to establish a clear correlation between cutting forces and tool condition, with particular emphasis on force components in the feed (X) and normal (Y) directions, which are known to be sensitive indicators of tool wear progression in milling operations. Figure 2 illustrates the experimental setup used for cutting force signal acquisition during the milling process.

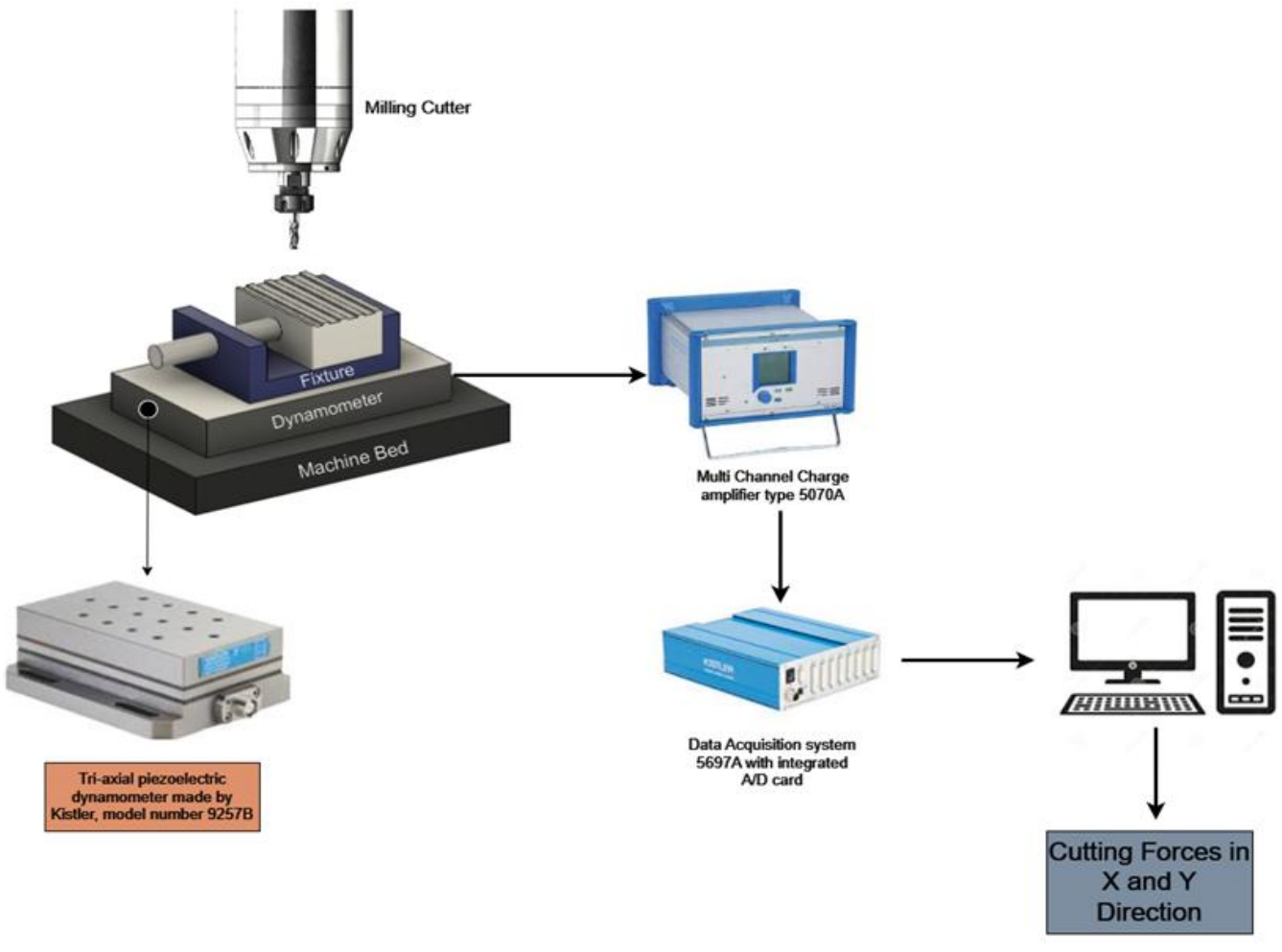

Figure 2. Schematic of experimental setup.

### 3.1. Machining operation and workpiece material

A slotting operation was performed on an Aluminum Alloy Al 6061 workpiece to estimate cutting forces in the feed and normal directions. Al 6061 was selected due to its widespread industrial use in aerospace, automotive, and general manufacturing applications, as well as its stable machinability and consistent mechanical properties. These characteristics allow repeatable machining conditions and minimize excessive tool–material interaction effects such as built-up edge formation, thereby enabling a clearer relationship between cutting force variations and tool wear progression. The use of Al 6061 is well suited for force-based Tool Condition Monitoring (TCM) studies, where consistent signal behavior is critical for reliable feature ex-traction and classification.

### 3.2. Cutting tool selection

A High-Speed Steel (HSS) end mill cutter was used to perform the machining operation. The technical specifications of the cutter are detailed in Table 2. HSS tools were selected due to their cost-effectiveness, widespread industrial usage, and predictable wear behavior compared to coated carbide or advanced tool materials. HSS tools exhibit gradual and progressive wear patterns, which are advantageous for controlled tool wear studies and facilitate clearer differentiation between tool condition states. Additionally, the use of a re-sharpenable HSS cutter aligns with practical manufacturing scenarios and enhances the applicability of the proposed TCM framework.

### 3.3. Selection of cutting parameters

The selection of cutting parameter ranges in this study was guided by both practical milling considerations and ranges commonly reported in force-based TCM literature. Previous TCM studies in milling have typically employed spindle speeds in the range of approximately 1000–3000 rpm, depth of cut values between 0.5–3 mm, and moderate feed rates to ensure stable cutting while enabling progressive tool wear development. The parameter ranges adopted in this work fall well within these commonly reported limits, ensuring comparability with existing studies while maintaining experimental safety and repeatability. In this study, spindle speeds of 1500–1700 rpm were selected to represent a moderate cutting speed regime suitable for HSS tools machining aluminum alloys. Operating within this range avoids excessive thermal loading that may lead to abrupt tool failure, while still generating sufficient cutting forces to reflect changes in tool condition. Lower spindle speeds may result in limited force sensitivity to wear, whereas significantly higher speeds can accelerate wear mechanisms beyond controlled monitoring conditions. The chosen spindle speed range therefore provides a balance between force signal stability and wear progression sensitivity. The depth of cut was varied between 0.75 mm and 2.25 mm to systematically increase cutting load and force magnitude across experiments. Depth of cut has a direct and proportional influence on cutting forces and is one of the most sensitive parameters affecting tool wear progression in milling operations. Selecting this range enables clear differentiation between light, moderate, and relatively heavy cutting conditions without inducing chatter or excessive tool deflection. Depths below this range may not produce sufficiently distinguishable force patterns for effective tool condition classification, while larger depths may compromise tool integrity, particularly for HSS cutters. Feed rates were selected in conjunction with spindle speed and depth of cut to maintain stable chip formation and consistent tool–workpiece interaction. By positioning the selected parameter ranges within those commonly adopted in prior TCM studies, the proposed experimental setup ensures both scientific validity and comparability with existing literature, while supporting reliable force-based tool condition monitoring.

Table 2. Specifications of HSS cutter.

| Parameter | Value |
|---|---|
| Length | 65 mm |
| Overhang | 40 mm |
| Number of Inserts | 4 |
| Diameter | 12 mm |
| Helix Angle | 35° |
| Type | Re-sharpenable |

### 3.4. Force measurement and data acquisition

Cutting force signals were acquired using a Kistler piezoelectric dynamometer (Type 9257B), which is designed to accurately measure cutting forces in the X, Y, and Z directions during machining operations. The specifications of the dynamometer, including sensitivity, measuring range, operating temperature, and natural frequency, are summarized in Table 3.

Table 3. Dynamometer specifications.

| Specification | Value |
|---|---|
| Maximum measuring range | 10 kN |
| Sensitivity | Fx, Fy: -7.5 pC/N<br>Fz: -3.7 pC/N |
| Operating Temperature | 0-70°C |
| Insulation Resistance | $>10^{13}$ Ω |
| Natural Frequency Fn (x,y) | ≈2.3 kHz |
| Capacitance (Fx, Fy, Fz) | ≈220 pF |
| Weight | 7.3 kg |
| Clamping area | 100 mm x 170 mm |

A sampling frequency of 11.6 kHz was selected to adequately capture the dynamic force variations associated with intermittent milling operations. The force signals generated by the dynamometer were transmitted through a Type 5070A charge amplifier before entering the 5697A data acquisition (DAQ) system equipped with an A/D card. The DAQ system was configured with appropriate sampling rate, measurement range, and channel settings to ensure accurate and reliable signal capture. The acquired data was visualized and stored for further processing and analysis.

### 3.5. Tool condition labeling and calibration

Tool condition labeling in this study was performed based on controlled machining progression and visual inspection of the cutting tool, following commonly adopted practices in milling tool condition monitoring. Tool conditions were categorized into distinct wear states corresponding to the progression of tool degradation during machining. Although international standards such as ISO 8688 provide guidelines for quantitative tool life evaluation, direct flank wear measurements were not continuously available during force signal acquisition. Therefore, tool condition calibration was performed using indirect indicators such as machining progression and observable wear characteristics. The experiments were conducted under controlled conditions using the selected parameters, and the collected dataset comprised cutting force signals along with corresponding spindle speed, depth of cut, and feed rate information. Subsequent sections of this paper present the analysis and results derived from this dataset, highlighting the relationship between force signal characteristics and tool condition during slotting operations.

## 4. Data cleansing

After analyzing the signals for force in the X direction for all experiments, the outliers – the data points that deviate significantly from the trend- were identified visually and removed from every plot to get better results. The outliers, if left unaddressed, have an impact on the statistical features of the entire data and can lead to erroneous conclusions. Fig. 3 displays a scatter plot depicting a typical sample of around 10000 data points of the cleaned dataset.

A similar process was conducted for the data of force in Y direction. This method of eliminating errors and inconsistencies was done to ensure accurate and reliable model training as explained further in the paper.

## 5. Machine learning approach

In this section, feature extraction, feature selection, classification through the K-Nearest Neighbours (KNN) algorithm has been elaborated and a model-agnostic approach is also introduced to provide insights on how predictions are made by the model.

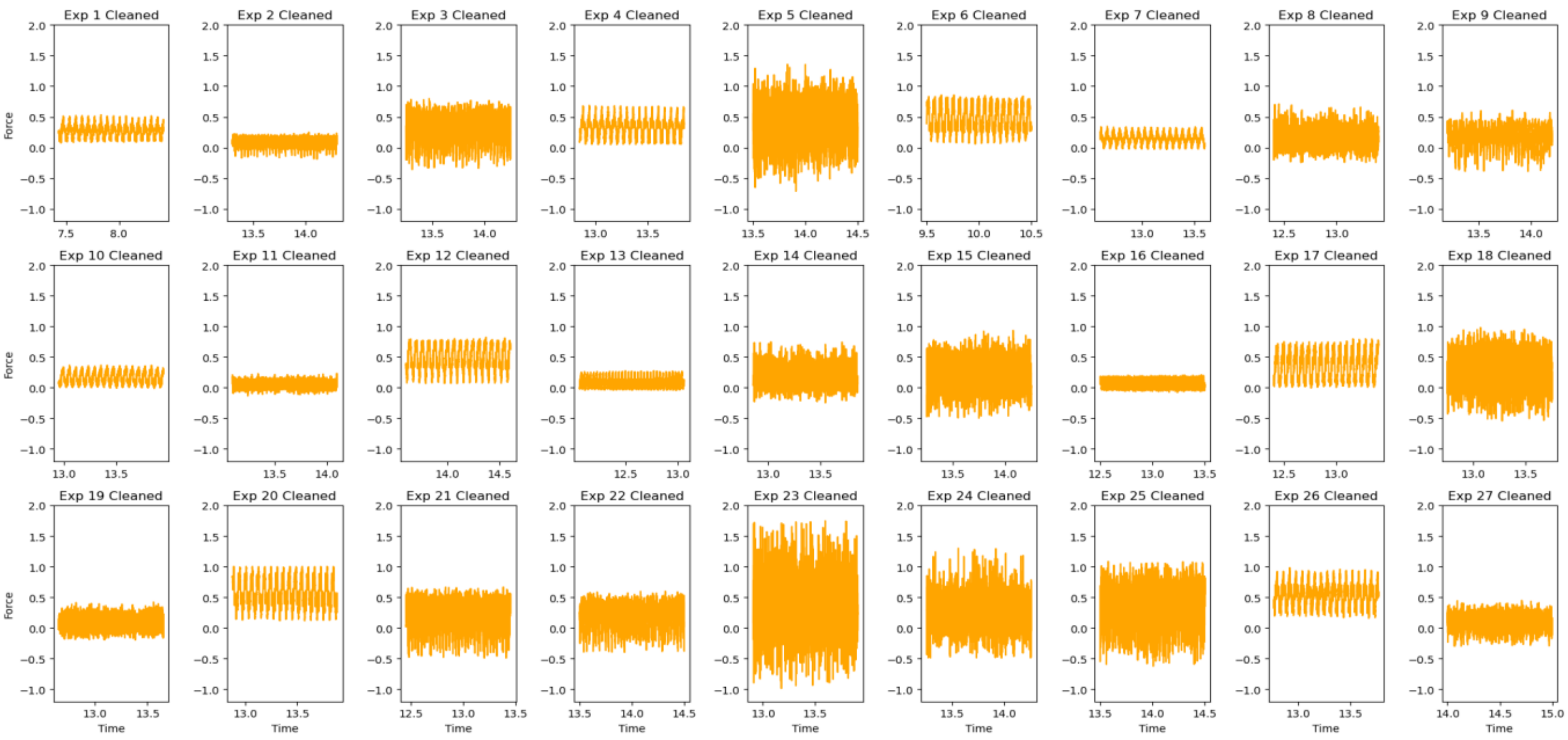

Figure 3. Scatter plots for 10000 data points of the cleaned data.

### 5.1. Feature extraction

Feature extraction is used for the transformation of raw data into numerical features which can be processed efficiently and also, simultaneously; preserve the information in the original dataset. Feature extraction was performed on the cleaned data for all the experiments. A total of 12 statistical features like (1) mean, (2) median, (3) kurtosis, (4) skewness, (5) standard error, (6) variance, (7) maximum, (8) minimum, (9) range, (10) summation, (11) standard deviation, and (12) standard error were extracted for force in X and Y direction.

### 5.2. Feature selection

Improving the prediction power of the models is essential and for that feature selection is a decisive step. It eliminates the redundant ones by selecting the most critical attributes that contribute the most in predicting the target variables. Feature selection improves the model performance, reduces computational complexity, mitigates high-dimensional data, and reduces the risk of over-fitting. To evaluate the significance of individual features, Decision tree classifier was used. The algorithm starts by evaluating different input features and all possible split points to determine which split results in the largest reduction in impurity, measured in terms of Gini impurity [24].

Further, while building the tree, the algorithm keeps track of how much each feature contributes to reducing impurity. This process continues recursively by selecting the best feature for splitting and adding up the importance score that shows how much one feature contributes towards the model's performance [25]. Finally, features with higher importance value scores are selected. The Decision Tree Classifier selected the top 10 features (minimum, skewness, median, summation, mean, kurtosis, standard error, range, maximum and mode), and hence the other two were removed.

### 5.3. Data augmentation strategy

Data augmentation was employed in this study to enhance the robustness of the classification model and to reduce Type II errors, which are particularly critical in Tool Condition Monitoring (TCM) applications. Due to practical constraints in machining experiments, the number of samples available for certain tool wear conditions is inherently limited, leading to sparse feature-space representation and increased misclassification risk. In this work, data augmentation was performed after statistical feature extraction from the cutting force signals. The augmentation was carried out in the feature space, rather than directly on the raw time-domain signals. For a given feature vector

$$x = [x_1, x_2, \dots, x_n], \tag{1}$$

augmented samples were generated by introducing small, controlled perturbations to the original feature values:

$$x' = x + \epsilon, \quad (2)$$

where $\boldsymbol{\epsilon}$ is a zero-mean perturbation vector whose magnitude is bounded by the local statistical dispersion (e.g., standard deviation) of each feature within the same tool condition class. This approach can be interpreted as a form of feature-level jittering, ensuring that the augmented samples remain physically meaningful and representative of the original class.

Unlike signal-level augmentation methods such as segment duplication or time shifting, the adopted feature-level augmentation preserves the global statistical characteristics of the force signals while increasing sample density in the feature space. The augmented data therefore differ from the original data only by small variations within the observed feature ranges, without introducing artificial patterns or extrapolating beyond the original data distribution. To control the risk of overfitting, several safeguards were applied. First, the magnitude of perturbation was strictly limited to prevent excessive deviation from the original feature vectors. Second, augmentation was applied uniformly across tool condition classes to avoid introducing class imbalance. Third, model performance was validated using train–test splitting and cross-validation, ensuring that the classifier generalized well to unseen data rather than memorizing augmented samples. For instance-based classifiers such as KNN, classification reliability depends strongly on the availability of representative neighboring samples. By increasing local sample density—particularly near class boundaries associated with transitional tool wear states—the augmented dataset improves neighborhood consistency and reduced ambiguity in distance-based decision making.

### 5.4. Model-agnostic white-box approaches in TCM

Interpretability of TCM systems is crucial, as their output directly influences maintenance scheduling, production continuity, and operational safety. Existing interpretable approaches in TCM are predominantly model-specific white-box methods, such as decision trees, rule-based systems, and interpretable support vector machine (SVM) formulations. In these approaches, transparency is derived from the internal structure of the learning algorithm itself, making the explanation inherently tied to a specific model architecture. While model-specific white-box models offer clear insights into decision rules, they suffer from limited flexibility. Any change in the underlying classifier requires redesigning the interpretability mechanism, and their performance can degrade when dealing with highly nonlinear and overlapping feature spaces commonly observed in milling force signals. Furthermore, model-specific explanations are often difficult to generalize across different operating conditions or machining setups. In contrast, a model-agnostic interpretability approach decouples the explanation mechanism from the predictive model. This allows interpretability tools to be applied independently of the learning algorithm, enabling the same explanation framework to be reused across different classifiers. Model-agnostic methods provide both global interpretability, which reveals overall feature influence trends, and local interpretability, which explains individual predictions—an essential requirement in safety-critical TCM applications. To further highlight the contribution of the proposed approach, a comparison with commonly used white-box models in TCM is presented in Table 4.

Table 4. Comparison of white-box approaches.

| Aspect | Model-Specific | Model-Agnostic |
|---|---|---|
| Interpretability source | Internal model structure | External explanation methods |
| Flexibility across models | Low | High |
| Local explanation | Limited | Strong (instance-level) |
| Global explanation | Moderate | Strong |
| Adaptability to new models | Requires redesign | Directly reusable |
| Suitability | Moderate | High |

Although model-agnostic interpretability techniques have been widely explored in general machine learning literature, their systematic application in force-based TCM for milling operations remains limited. In particular, the integration of a model-agnostic explanation framework with an instance-based classifier such as KNN has not been adequately investigated for understanding tool wear progression and decision rationale in milling processes. In this work, a KNN-based model-agnostic white-box framework is proposed, where interpretability is achieved using external explanation techniques rather than the internal mechanics of the classifier. This enables detailed insight into how different statistical force features influence tool condition classification, while retaining the flexibility to replace the underlying model without sacrificing interpretability. This approach bridges the gap between high predictive performance and actionable transparency in TCM systems.

### 5.5. Mathematical formulation of model-agnostic interpretability and LIME

The proposed model-agnostic white-box framework treats the trained classifier as a black-box prediction function that maps an input feature vector to a tool condition class. Let the trained model be represented as:

$$\hat{y} = f(x), \boldsymbol{x} \in \mathbb{R}^{n} \tag{3}$$

where $\boldsymbol{x}$ denotes the extracted statistical force features and $\hat{y}$ represents the predicted tool condition. In a model-agnostic setting, interpretability does not rely on the internal parameters or structure of $f(\cdot)$. Instead, explanations are generated by analyzing how variations in the input features influence the model output.

This framework enables interpretability independent of the underlying classifier and remains valid even if the prediction model is replaced. The approach is particularly suited to TCM applications, where model updates may be required due to changes in machining conditions, tools, or materials. To achieve instance-level interpretability, the Local Interpretable Model-Agnostic Explanations (LIME) method is employed. LIME explains individual predictions by approximating the behavior of the complex model $f(\cdot)$ in the vicinity of a specific data instance $\mathbf{x}_0$ using a simple and interpretable surrogate model $g(\cdot)$.

The objective of LIME is defined as:

$$arg\,\min_{g} L\left(f, g, \pi_{x_0}\right) + \Omega(g) \tag{4}$$

where $L\left(f, g, \pi_{\mathrm{x}_0}\right)$ measures the fidelity of the surrogate model $g$ to the original model f in the locality of $\boldsymbol{x}_0$, $\pi_{x_0}$ defines a proximity measure around $x^0$, and $\Omega(g)$ represents a complexity penalty that enforces interpretability.

For tabular data, the surrogate model g is typically a sparse linear model:

$$g(z) = w^0 + \sum_{i=1}^{n} w_{\mathrm{i}}\, z_{\mathrm{i}} \tag{5}$$

where $z$ represents the perturbed samples around $x_0$ and $w_{\mathrm{i}}$ indicates the contribution of the i-th feature to the local prediction. Features with larger absolute values of $w_i$ have higher influence on the classification decision, providing instance-level interpretability.

When combined with the instance-based KNN classifier, this formulation provides physically meaningful explanations by linking force-feature variations to local neighborhood-based decisions. As a result, the proposed framework delivers transparent, instance-level reasoning while preserving the flexibility and predictive strength required for real-world Tool Condition Monitoring systems.

## 6. Design of KNN model

The selection of an appropriate classification algorithm for TCM depends strongly on the characteristics of the machining process, sensor data, and practical deployment requirements. Milling operations generate highly nonlinear, noisy, and overlapping force signals due to inter-mittent cutting, varying chip load, and tool–workpiece interactions. These characteristics im-pose specific constraints on the choice of learning algorithm.

K-Nearest Neighbors (KNN) was selected in this study due to its strong adaptability to the scenario-specific characteristics of milling-based TCM. Unlike parametric models such as SVMs or neural networks, KNN is a non-parametric, instance-based learning algorithm that does not assume any predefined decision boundary or underlying data distribution [26]. This property is particularly advantageous for milling force data, where feature distributions often overlap across tool wear states and evolve with changing cutting conditions. Compared to deep learning ap-proaches listed in Table 1, KNN requires significantly fewer training samples and does not rely on extensive hyperparameter tuning or computational resources. This makes it well suited for industrial TCM applications, where labeled data is limited and real-time deployment is required. Additionally, KNN inherently supports incremental learning, allowing new machining data to be incorporated without retraining the entire model—an important requirement for adaptive tool monitoring systems. From an interpretability standpoint, KNN aligns naturally with the objec-tives of this study. Each classification decision is based on the proximity of a test instance to neighboring samples in the feature space, providing an intuitive physical interpretation that directly relates to historical tool behavior. When combined with model-agnostic explanation techniques such as LIME, KNN enables both global understanding of feature influence and local, instance-level reasoning for individual tool condition predictions. Furthermore, unlike tree-based models that may oversimplify nonlinear relationships or SVM-based approaches that depend heavily on kernel selection, KNN preserves local data structure and responds effectively to subtle changes in force features caused by progressive tool wear. This makes KNN particu-larly suitable for distinguishing between closely related wear states such as "initial wear" and "bad" tool conditions observed in milling operations. Therefore, the choice of KNN in this work is motivated not only by its proven performance in fault diagnosis literature but also by its strong compatibility with the data characteristics, interpretability requirements, and practical con-straints of milling tool condition monitoring.

### 6.1. Model training

The data was trained by using KNN model with standard hyperparameters (termed as Vanilla KNN model) and then altering the hyperparameters (termed as tuned KNN model). Thus, the Vanilla KNN model uses the default hyperparameters without any modifications. Tuned KNN model involves optimization of hyperparameters to achieve better performance. These two models were used to establish baseline performance and understand the impact of parameter tuning on the model's performance. The model's performance was corroborated by using the cross-validation technique [27]. K-fold cross-validation was used in which the classification was performed on the training data for K between 5 and 10.

#### 6.1.1. Design of KNN model considering standard hyperparameters (Vanilla)

The entire data was used as training data to train the model. Default parameters of the KNN model were used. Training was done for both force signals in X and force signals in Y direction to understand the dominance of one model over the other.

a) Model's performance on force signals in X direction

The model was trained using the dataset for Force in X-Direction. The training was done for both raw and augmented signals.

i. Raw signals

A total of 2726 samples were used for training the data and the confusion matrix was plotted as shown in Fig. 4. Even though the model performed well, a significant amount of type II error was observed. Fig. 5 represents a visualization of how the KNN model performs on raw data of force in the X direction. To reduce the dimensionality of the model, Principal component analysis was used, and the dimensionality was reduced from 10 to 2. It illustrates the training data represented by three output classes with three different colors. A total of 5 test instances were considered, and the model classified them correctly, as justified by the selection of its 4 neighbors from the diagram. Thus, the model performed well for force in the X direction.

Tool health based on raw force X signal data

| Actual \ Predicted | Good Condition | Initial Wear | Progressed Wear | SUM |
| --- | --- | --- | --- | --- |
| Good Condition | 866<br>31.77% | 24<br>0.88% | 18<br>0.66% | 908<br>95.37%<br>4.63% |
| Initial Wear | 50<br>1.83% | 831<br>30.48% | 28<br>1.03% | 909<br>91.42%<br>8.58% |
| Progressed Wea | 42<br>1.54% | 23<br>0.84% | 844<br>30.96% | 909<br>92.85%<br>7.15% |
| SUM | 958<br>90.40%<br>9.60% | 878<br>94.65%<br>5.35% | 890<br>94.83%<br>5.17% | 2541 / 2726<br>93.21%<br>6.79% |

Figure 4. Confusion matrix of raw data for force signals in X direction.

ii. Augmented signals

The raw data was augmented to deal with Type II error and thereby improve performance. A total of 5426 samples were used for training the model. Augmenting the data led to significant decrease in Type II error. Since this is a real-world application, it is necessary to remove false positives. Fig. 6 shows confusion matrix of augmented data for force signals in X direction.

b) Model's performance on force signals in Y-Direction

The model was trained using the dataset for Force in Y-Direction. The training was done for both Raw and Augmented signals.

i. Raw signals

A total of 2726 samples were used for training, and the confusion matrix was plotted as shown below. Fig. 7 confusion matrix of raw data for force signals in Y direction. In case of the data of Force signals in Y direction, the KNN classifier model does not perform well. This is illustrated in Fig. 8 where the neighbors for the test instances are not what were expected, leading to misclassification and thus lesser accuracy in the data. This also leads to an increase in Type II error.

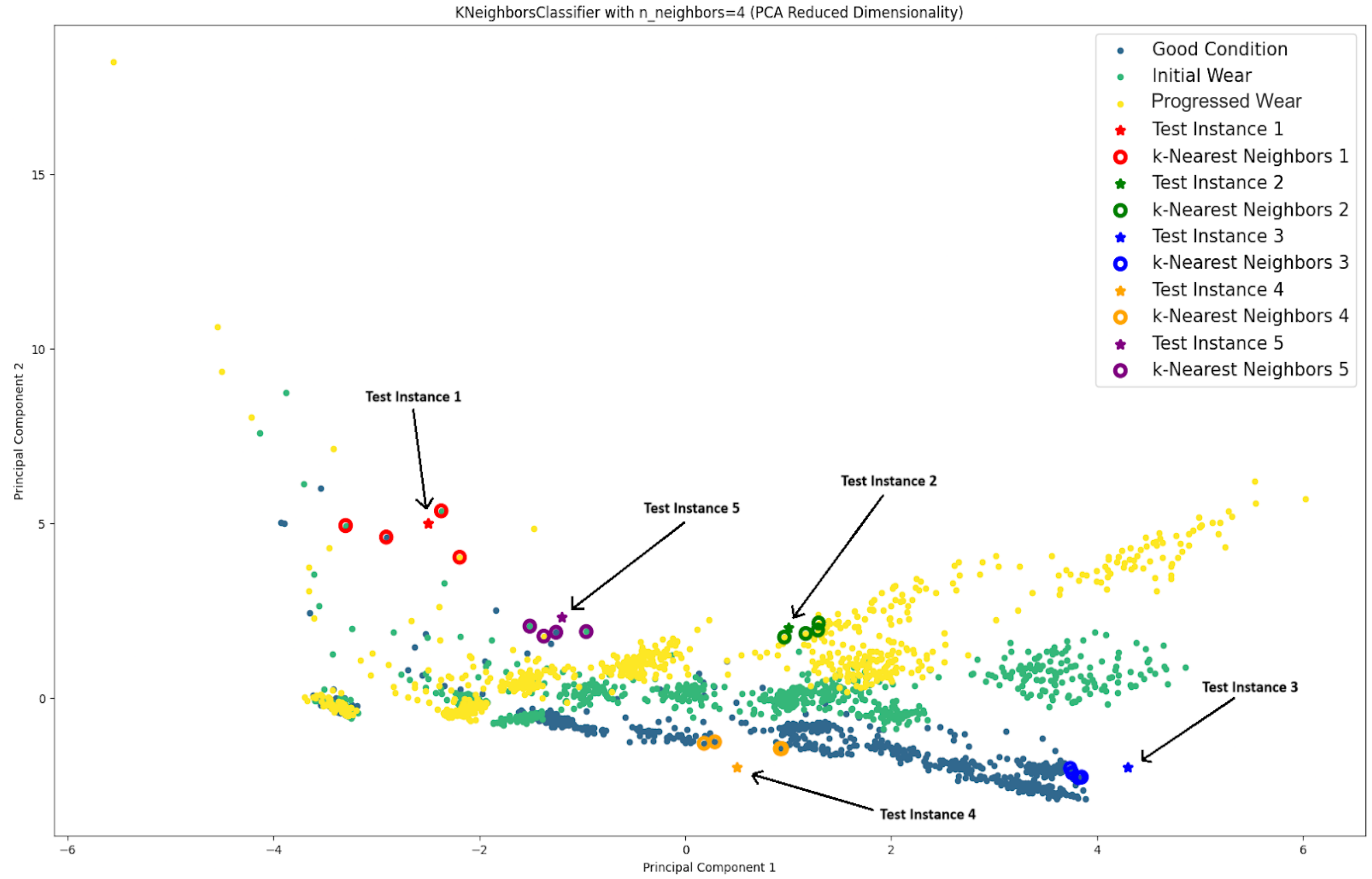

Figure 5. Scatter plot of raw data for force signals in X direction.

Tool health based on augmented force X signal data

| Actual \ Predicted | Good Condition | Initial Wear | Progressed Wear | SUM |
| --- | --- | --- | --- | --- |
| Good Condition | 1766<br>32.55% | 30<br>0.55% | 12<br>0.22% | 1808<br>97.68%<br>2.32% |
| Initial Wear | 4<br>0.07% | 1694<br>31.22% | 111<br>2.05% | 1809<br>93.64%<br>6.36% |
| Progressed Wear | 4<br>0.07% | 53<br>0.98% | 1752<br>32.29% | 1809<br>96.85%<br>3.15% |
| SUM | 1774<br>99.55%<br>0.45% | 1777<br>95.33%<br>4.67% | 1875<br>93.44%<br>6.56% | 5212 / 5426<br>96.06%<br>3.94% |

Figure 6. Confusion matrix of augmented data for force signals in X direction.

Tool health based on raw force Y signal data

| Actual \ Predicted | Good Condition | Initial Wear | Progressed Wear | SUM |
| --- | --- | --- | --- | --- |
| Good Condition | 696<br>25.53% | 116<br>4.26% | 96<br>3.52% | 908<br>76.65%<br>23.35% |
| Initial Wear | 202<br>7.41% | 514<br>18.86% | 193<br>7.08% | 909<br>56.55%<br>43.45% |
| Progressed Wear | 99<br>3.63% | 208<br>7.63% | 602<br>22.08% | 909<br>66.23%<br>33.77% |
| SUM | 997<br>69.81%<br>30.19% | 838<br>61.34%<br>38.66% | 891<br>67.56%<br>32.44% | 1812 / 2726<br>66.47%<br>33.53% |

Figure 7. Confusion matrix of raw data for force signals in Y direction.

ii. Augmented signals

The raw data was augmented to check if the model's performance has improved. A total of 5422 samples were used for training the model. Fig. 9 shows confusion matrix of augmented data for force signals in Y direction.

The model used for the augmented data of Force signals in the X direction outperformed the corresponding model for Force signals in the Y direction giving the highest training accuracy of 96% as shown in Table 5.

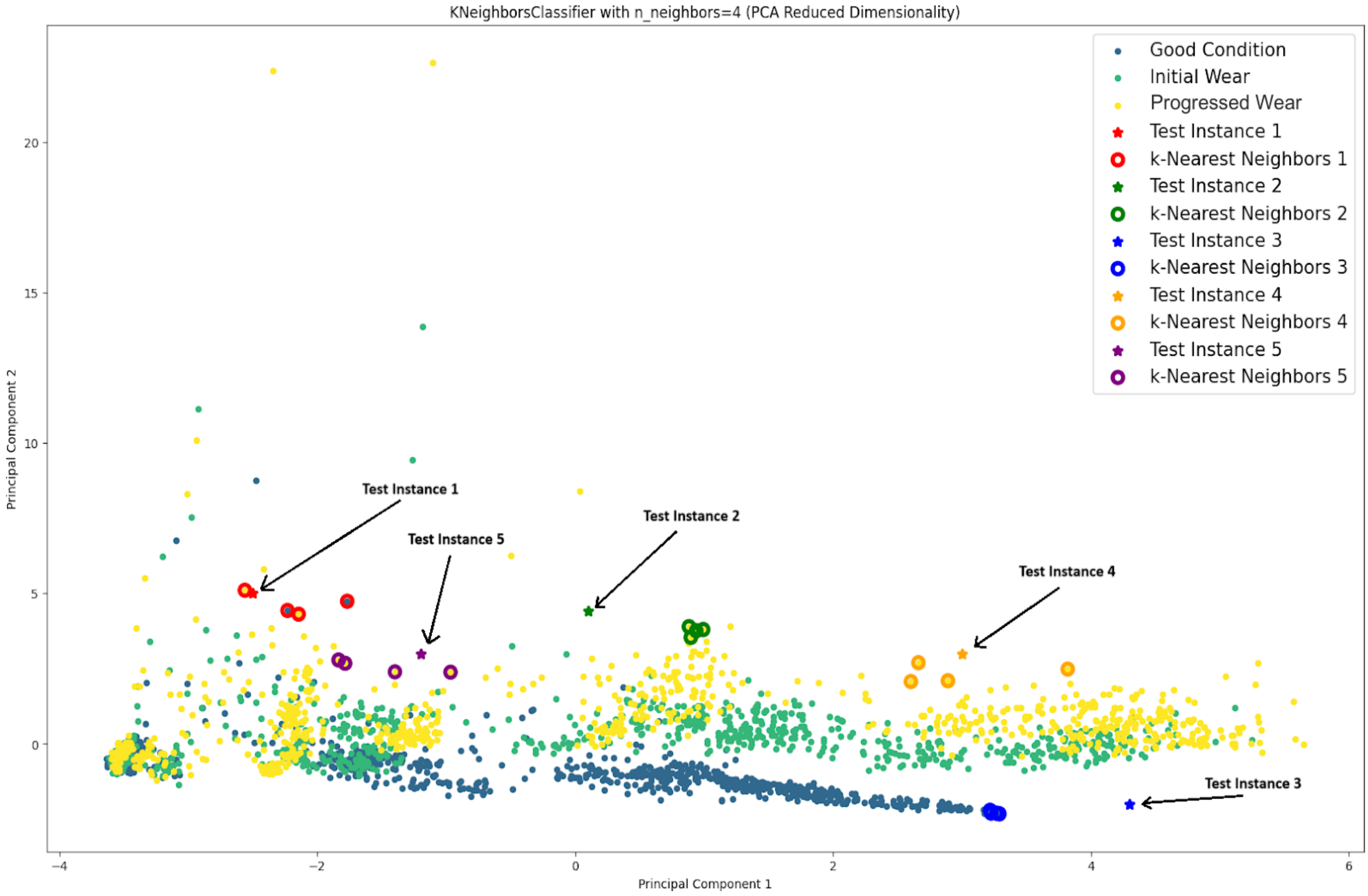

Figure 8. Scatter plot of raw data for force signals in Y direction.

| Tool health based on raw force Y signal data | | | | |
|---|---|---|---|---|
| Predicted / Actual | Good Condition | Initial Wear | Progressed Wear | SUM |
| Good Condition | 696<br>25.53% | 116<br>4.26% | 96<br>3.52% | 908<br>76.65%<br>23.35% |
| Initial Wear | 202<br>7.41% | 514<br>18.86% | 193<br>7.08% | 909<br>56.55%<br>43.45% |
| Progressed Wear | 99<br>3.63% | 208<br>7.63% | 602<br>22.08% | 909<br>66.23%<br>33.77% |
| SUM | 997<br>69.81%<br>30.19% | 838<br>61.34%<br>38.66% | 891<br>67.56%<br>32.44% | 1812 / 2726<br>66.47%<br>33.53% |

Figure 9. Confusion matrix of augmented data for force signals in Y direction.

Table 5. KNN Models' accuracies.

| **Accuracy for force signals in X-Direction** | | **Accuracy for force signals in Y-Direction** | |
|---|---|---|---|
| Raw data | Augmented data | Raw data | Augmented data |
| 92% | 96% | 68% | 78% |

Table 6 shows how significantly the type 2 error decreased after augmentation of data. The type 2 error was the lowest for Force in X-Direction at 0.14%. Thus, the Force signals in X-Direction were substantially more dominant than Force signals in Y-Direction and are used for further inferences.

Table 6. Type 2 Percentage error.

| **Type 2 error for force signals in X-Direction** | | **Type 2 error for force signals in Y-Direction** | |
|---|---|---|---|
| Raw data | Augmented data | Raw data | Augmented data |
| 3.04 | 0.14 | 11.04 | 6.41 |

The classification performance obtained using X-directional force signals, compared to Y-directional signals, can be explained by the underlying mechanics of the milling and slotting process employed in this study. During slotting operations, the feed direction (X-direction) experiences direct and continuous interaction between the cutting tool and the workpiece, making the corresponding force component responsive to changes in tool condition. As tool wear progresses, phenomena such as edge rounding, flank wear, and increased friction at the tool–workpiece interface influence the resistance encountered along the feed direction. This results in noticeable variations in X-directional cutting forces, which are reflected in the extracted statistical features. Consequently, changes in tool condition tend to produce improved separability between wear states in the X-direction feature space. In contrast, the Y-direction force component primarily reflects lateral and normal forces arising from radial engagement, tool runout, and variations in cutter immersion. These forces are more affected by process-related factors such as vibration, machine-tool compliance, and intermittent tooth engagement, particularly in milling operations. As a result, Y-direction force signals generally exhibit higher variability and greater overlap across different tool wear conditions, which can reduce their discriminative capability for tool condition classification. Furthermore, the intermittent nature of milling causes the normal force component to fluctuate due to periodic entry and exit of cutter teeth, making it less responsive to gradual wear progression compared to the feed-direction force. This increased variability contributes to reduced feature stability and to the comparatively lower classification accuracy and higher Type II error observed for Y-direction force-based models.

Beyond these general cutting mechanics, the observed behavior of X-directional force signals can be further explained through engineering considerations related to tool geometry, cutting kinematics, and load transfer during the slotting operation. In the experimental setup, material removal primarily occurs along the feed direction, which aligns with the X-axis. As a result, cutting forces generated due to chip formation and frictional resistance are largely transferred along this direction, making the X-direction force component more representative of the cutting load experienced by the tool. The geometry of the HSS end mill with a helix angle of 35° also influences force distribution. The helical cutting edges introduce an oblique cutting action, causing the resultant cutting force to be decomposed into axial, radial, and tangential components. A larger helix angle promotes smoother chip evacuation and gradual engagement of the cutting edge with the workpiece, while redistributing a portion of the tangential cutting force along the feed direction. Consequently, variations in tool condition, such as flank wear or edge rounding, tend to have a more pronounced effect on the feed-direction force component.

During slotting operations, the cutter is fully immersed in the workpiece, resulting in relatively symmetric radial engagement. Under such conditions, lateral (Y-direction) force components generated by opposing cutter teeth may partially offset each other, particularly when tool runout is minimal. In contrast, the feed-direction force accumulates more consistently due to continuous resistance against material removal. This difference in load transfer helps explain why X-directional forces are generally more indicative of tool wear progression than Y-directional forces. Additionally, progressive tool wear increases friction at the tool–chip and tool–workpiece interfaces, which contributes to an increase in tangential and feed-direction forces. The Y-direction force, however, remains more influenced by transient effects such as vibration, cutter deflection, and machine-tool compliance, making it less stable and more variable. This variability can reduce the effectiveness of Y-direction force features in distinguishing between adjacent tool wear states. Overall, the combined effects of cutting directionality, helical tool geometry, force decomposition, and load transfer mechanisms provide a physical basis for the observed differences in classification performance between X- and Y-directional force signals.

#### 6.1.2. Design of KNN model considering hyperparameters tuning

Hyperparameters are parameters that are set before training the data and do not take part in the learning process, yet they play a crucial role in determining the performance and generalization capability of machine learning models. To identify the optimal hyperparameter configuration for the K-Nearest Neighbors (KNN) classifier, the GridSearchCV technique was employed. This method systematically evaluates different combinations of hyperparameters using cross-validation and selects the configuration that yields the best performance [28]. The grid search was conducted using a 5-fold cross-validation strategy, ensuring that model selection was not biased toward a particular data split. The hyperparameters considered in the grid search included the number of neighbors, distance metric, and weighting scheme, which are known to have a significant influence on KNN performance. The number of neighbors ($n_neighbors$) determines how many nearest data points in the training set are considered when predicting the class of a test instance. Instead of fixing this value a priori, $n_neighbors$ was varied over a predefined range to evaluate its impact on classification accuracy and Type II error. The optimal value identified through grid search was $n_neighbors = 4$. The distance metric defines how similarity between data points is computed. Commonly used metrics such as Minkowski, Euclidean, Manhattan, and Cosine distances were considered during tuning. Based on cross-validation results, the Manhattan distance (L1 norm) was selected as the best-performing metric. The Manhattan distance between two points $P_1(x_1, y_1)$ and $P_2(x_2, y_2)$in a two-dimensional space is defined as:

$$d(P_1, P_2) = | x_2 - x_1 | + | y_2 - y_1 | \quad (6)$$

This metric is particularly suitable for high-dimensional feature spaces, as it assigns equal importance to each dimension and is less sensitive to outliers compared to Euclidean distance [29]. The weighting scheme determines the contribution of each neighbor to the final prediction. During tuning, both uniform and distance-based weighting were evaluated. When distance-based weighting is applied, closer neighbors exert greater influence on the classification outcome. Mathematically, if $d_i$is the distance of the $i$-th nearest neighbor, its weight $w_i$is given by:

$$w_i = \frac{1}{d_i} \quad (7)$$

The predicted class is then obtained by computing the weighted sum of the class labels of the nearest neighbors. The grid search identified distance-based weighting as the optimal choice, as it improved local decision reliability and reduced misclassification near class boundaries. The complete hyperparameter search space explored using GridSearchCV is summarized in Table 7.

Table 7. Hyperparameter search space used in GridSearchCV.

| **Hyperparameter** | **Values used** |
|---|---|
| Number of neighbors ($n_neighbors$) | 1 – 15 |
| Distance metric | Euclidean, Manhattan, Cosine |
| Weight function | Uniform, Distance |
| Cross-validation folds | 5 |

Based on the grid search results, the best estimator was identified as a KNN model with $n_neighbors = 4$, Manhattan distance, and distance-based weighting. This tuned configuration consistently outperformed the vanilla KNN model and was therefore used for all subsequent analyses and reported results.

The data was split into training and testing. The model was trained with 90% of the data and tested with 10% of the data. This was done to check the model's ability to classify the unseen data. A Total of 543 samples were used for testing. Fig. 10 and 11 shows confusion matrix of test and train data respectively. As seen in the above confusion matrix from Fig. 10, the type II error was completely eliminated in the testing set after tuning the hyperparameters.

Tool health based on Test data

| Actual \ Predicted | Good Condition | Initial Wear | Progressed Wear | SUM |
|---|---|---|---|---|
| Good Condition | 184<br>33.89% | 2<br>0.37% | 1<br>0.18% | 187<br>98.40%<br>1.60% |
| Initial Wear | 0<br>0.00% | 164<br>30.20% | 6<br>1.10% | 170<br>96.47%<br>3.53% |
| Progressed Wear | 0<br>0.00% | 9<br>1.66% | 177<br>32.60% | 186<br>95.16%<br>4.84% |
| SUM | 184<br>100.00%<br>0.00% | 175<br>93.71%<br>6.29% | 184<br>96.20%<br>3.80% | 525 / 543<br>96.69%<br>3.31% |

Figure 10. Confusion matrix of test data.

Tool health based on Trained data

| Actual \ Predicted | Good Condition | Initial Wear | Progressed Wear | SUM |
|---|---|---|---|---|
| Good Condition | 1603<br>32.83% | 11<br>0.23% | 7<br>0.14% | 1621<br>98.89%<br>1.11% |
| Initial Wear | 3<br>0.06% | 1600<br>32.77% | 36<br>0.74% | 1639<br>97.62%<br>2.38% |
| Progressed Wear | 3<br>0.06% | 50<br>1.02% | 1570<br>32.15% | 1623<br>96.73%<br>3.27% |
| SUM | 1609<br>99.63%<br>0.37% | 1661<br>96.33%<br>3.67% | 1613<br>97.33%<br>2.67% | 4773 / 4883<br>97.75%<br>2.25% |

Figure 11. Confusion matrix of trained data.

Fig. 12 shows the representation for the KNeighboursClassifier Model for the tuned model. As can be interpreted from the image, the model accurately selects the neighbors for the test instances that help in making the classification. This gives us a visual representation of how KNN model works and how it makes the classification. The testing accuracy was found to be 95% and the training accuracy was 98%. The tool-condition wise accuracy is detailed in Table 8.

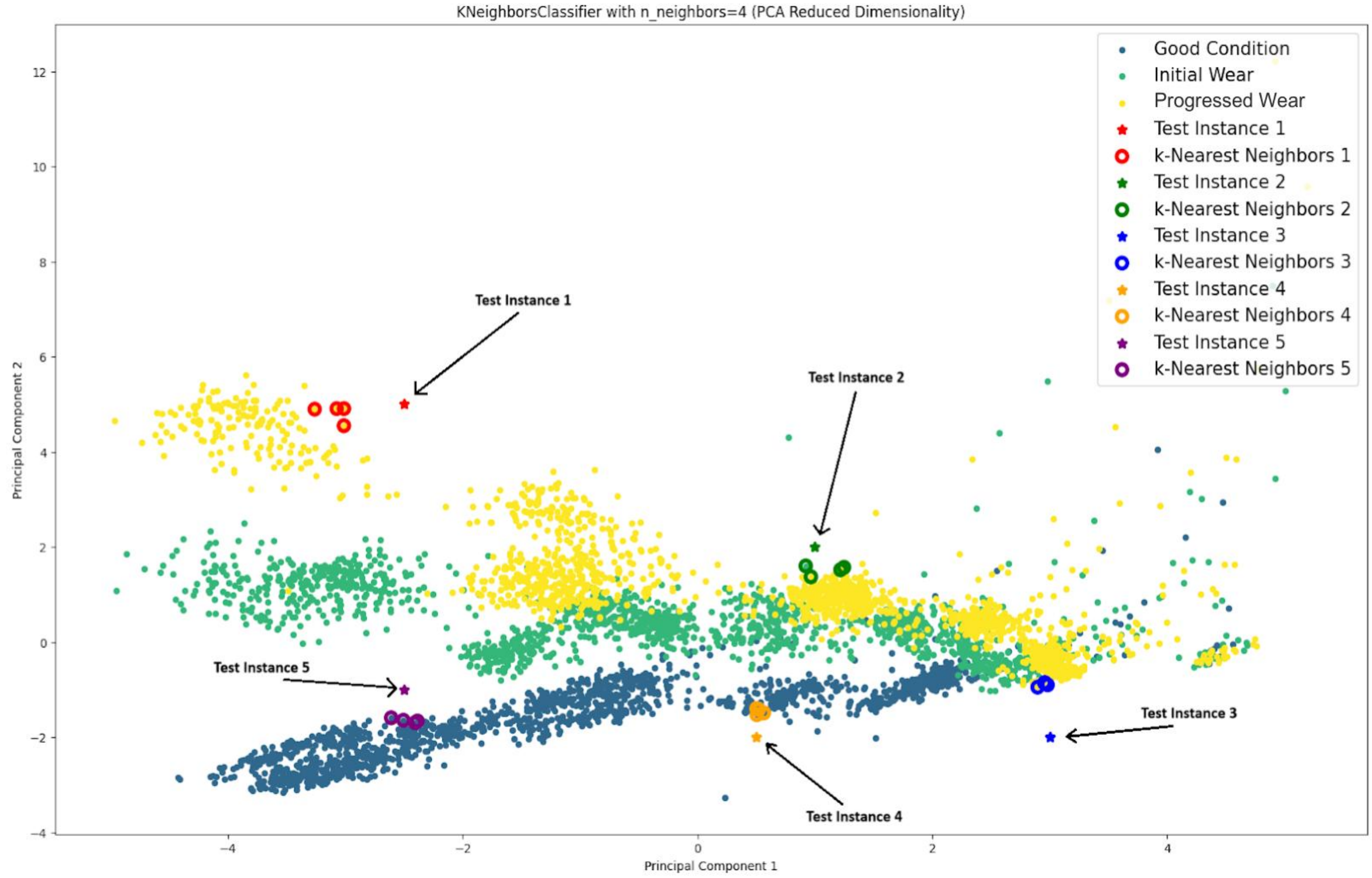

Figure 12. Scatter plot of the tuned model.

Table 8. Descriptive accuracy.

| Precision | Recall (TP) | F1-Score | ROC AUC | FP Rate | Condition |
|---|---|---|---|---|---|
| 0.99 | 0.99 | 0.99 | 0.99 | 0.01 | Good |
| 0.94 | 0.94 | 0.94 | 0.99 | 0.07 | Initial wear |
| 0.94 | 0.95 | 0.94 | 0.99 | 0.05 | Bad |

## 7. Results and discussion

This section discusses results after performing feature extraction, feature selection, and then classification through KNN. After consideration of the global representation for tuned KNN and vanilla KNN, and local representation for tuned KNN and vanilla KNN, the White-box results are presented. Entire data was trained using the Vanilla KNN model, and the model could classify 96% of the data correctly. After using the GridSearchCV technique for hyperparameter tuning, classification was performed again using the tuned model on the entire data. The model could classify 94% of the data points correctly. The reason for a decrease in accuracy was that the Vanilla KNN model was more prone to overfitting, whereas the tuned KNN model reduced this risk and created a more generalized model so that it would perform better on the unseen data. The Tuned model noticeably performed better than the Vanilla model. The Vanilla model had a testing accuracy of 87% which increased to 96% after tuning the hyperparameters.

### 7.1. Training and testing evaluation

Train-test splits were done on the entire data, and the performance of the model was analyzed over different percentages of the split, as shown in Fig. 13. The classification was done for the following optimal hyperparameters: n_neighbours = 4, metric=’manhattan’, weights = ‘distance’.

The classification results for the varying values of K-fold Cross Validation are as shown in Fig. 14.

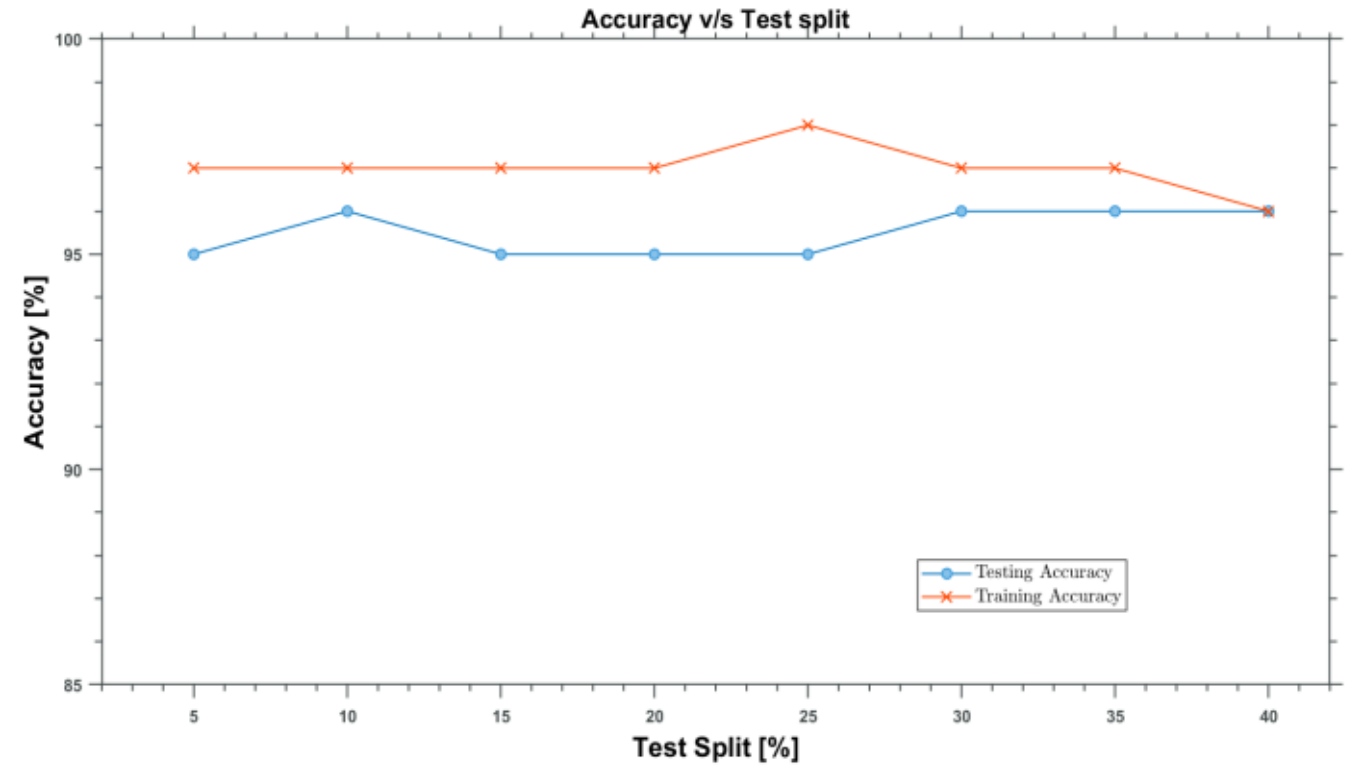

Figure 13. Training and testing accuracies vs test split.

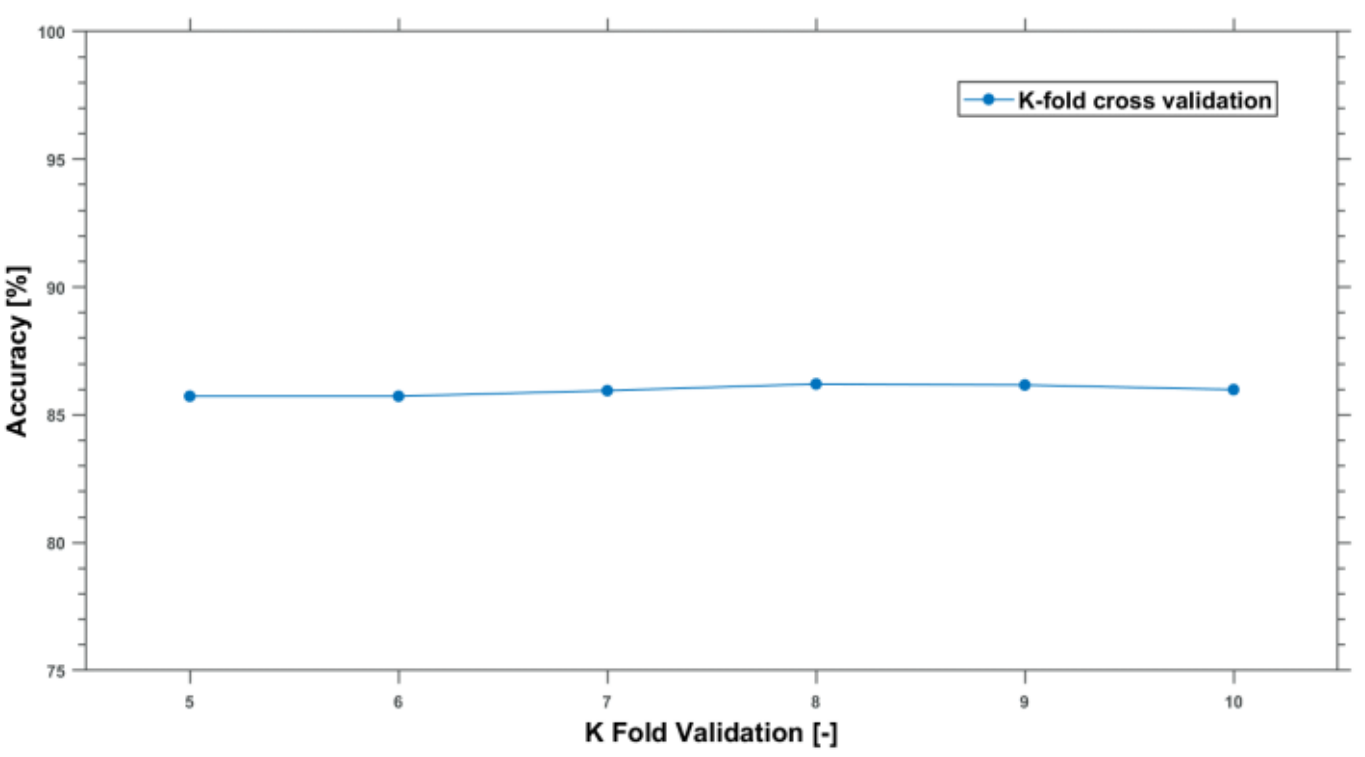

Figure 14. Training accuracy & vs K-fold.

### 7.2. White-box results

Global representation of KNN model refers to a holistic understanding of the input data. It captures the high-level features that contribute to the decision making of the model. Local representation of KNN model, on the other hand, focuses on specific, smaller details within the input data. It represents relevant features those make classification decisions for single instances [30].

#### 7.2.1. Global representation for KNN

Global representation is done to gain a high-level understanding of the data and identify the key trends as shown in Table 9 and Table 10.

For KNN with best parameters.

Table 9. Global Representation of Tuned KNN model.

| Weight | Feature |
|---|---|
| 0.0751 ± 0.0121 | Skewness |
| 0.0072 ± 0.0062 | Maximum |
| 0.0048 ± 0.0014 | Mode |
| 0.0042 ± 0.0036 | Standard Error |
| 0.0042 ± 0.0069 | Kurtosis |
| 0.0041 ± 0.0041 | Mean |
| 0.0029 ± 0.0014 | Median |
| 0.0028 ± 0.0052 | Summation |
| 0.0022 ± 0.0034 | Range |
| 0.0013 ± 0.0036 | Minimum |

For vanilla KNN

Table 10. Global representation of Vanilla KNN model.

| Weight | Feature |
|---|---|
| 0.1543 ± 0.0220 | Skewness |
| 0.1337 ± 0.0115 | Minimum |
| 0.1153 ± 0.0175 | Mean |
| 0.1077 ± 0.0169 | Summation |
| 0.0971 ± 0.0087 | Median |
| 0.0792 ± 0.0129 | Maximum |
| 0.0781 ± 0.0115 | Range |
| 0.0748 ± 0.0225 | Standard Error |
| 0.0659 ± 0.0100 | Mode |
| 0.0306 ± 0.0067 | Kurtosis |

In this representation, the "Weights" column indicates the weights associated with each feature, the "Feature" column lists the names of the features. This representation is helpful for understanding which features contribute more to a certain metric or model, and it can provide insights into feature selection, interpretation, and potentially optimization of the model. Decision trees help visualize the feature importance and provide insights into the decision-making process as represented in Fig. 15 and Fig. 16.

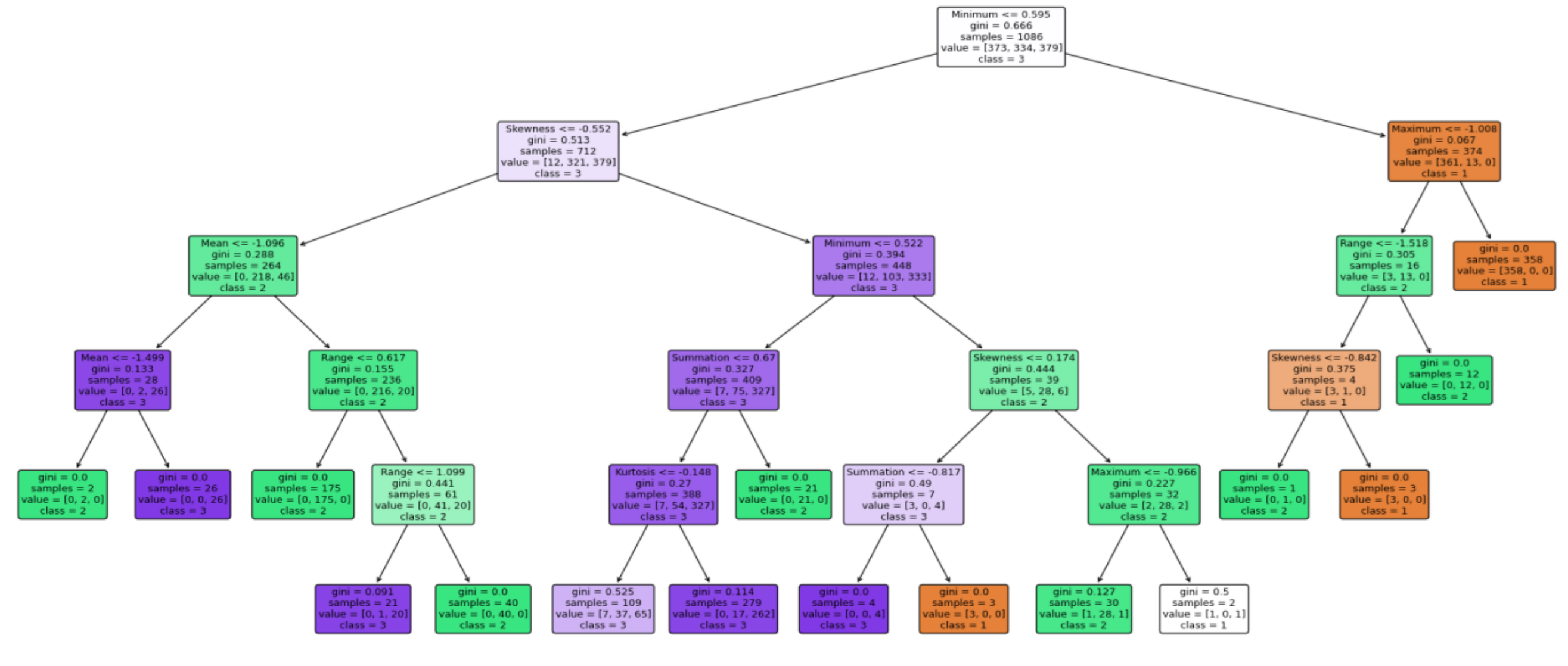

Figure 15. Decision tree of Tuned KNN model.

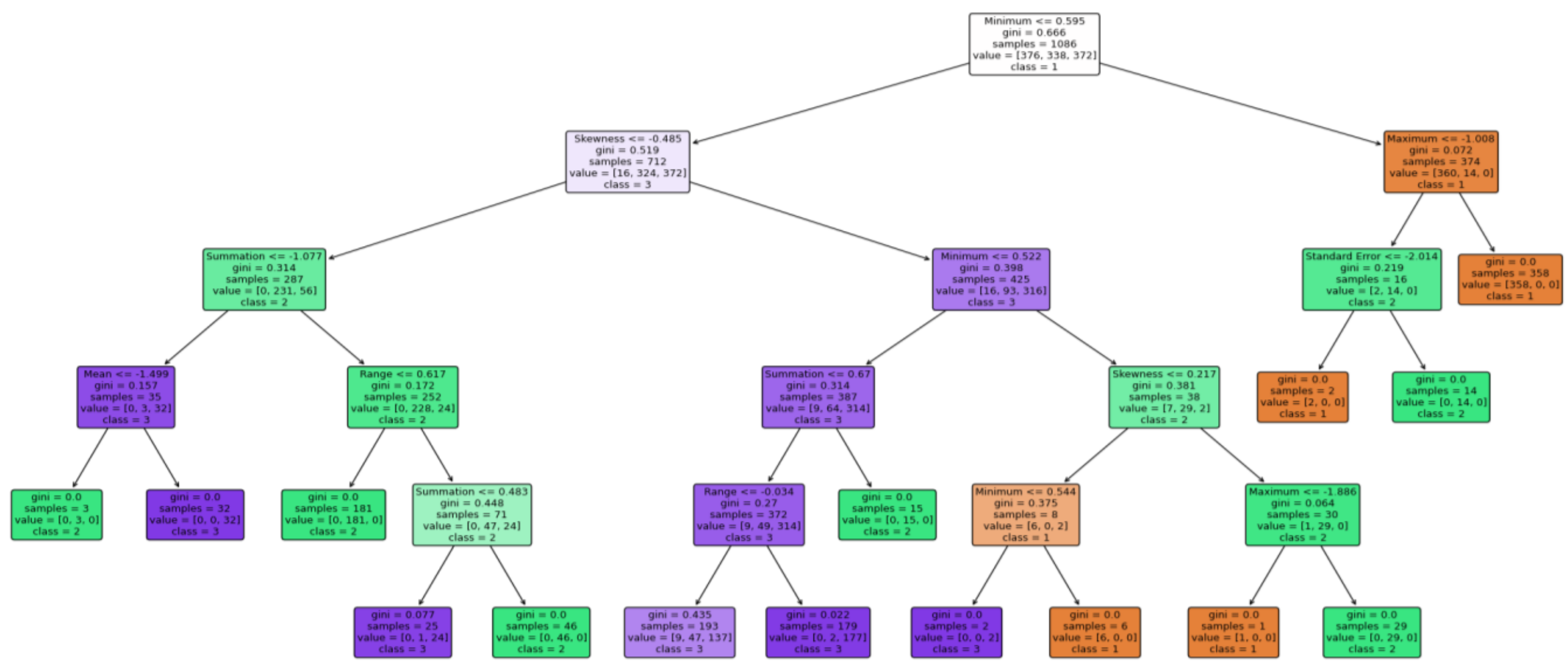

Figure 16. Decision tree of Vanilla KNN model.

### 7.2.2. Local representation

In this representation, the LIME (Local Interpretable Model-Agnostic Explanations) framework is used to enhance the interpretability of our classification model. The local representation of the k-NN classifier helps in explaining the decision-making process of the model for individual data points. This is crucial for establishing trust and comprehensibility in machine learning applications, especially when model outputs influence real-world decisions as in our case.

For KNN with best parameters

By leveraging LIME's capability to approximate local behavior around a chosen data instance, we were able to highlight the features' contributions that influenced the k-NN model's prediction for different classes. In Fig. 17 and Fig. 18, darker shades of green represent high positive influence of a parameter on the decision of the classification model, while darker shades of red represents high negative influence of a parameter on the decision of the classification model.

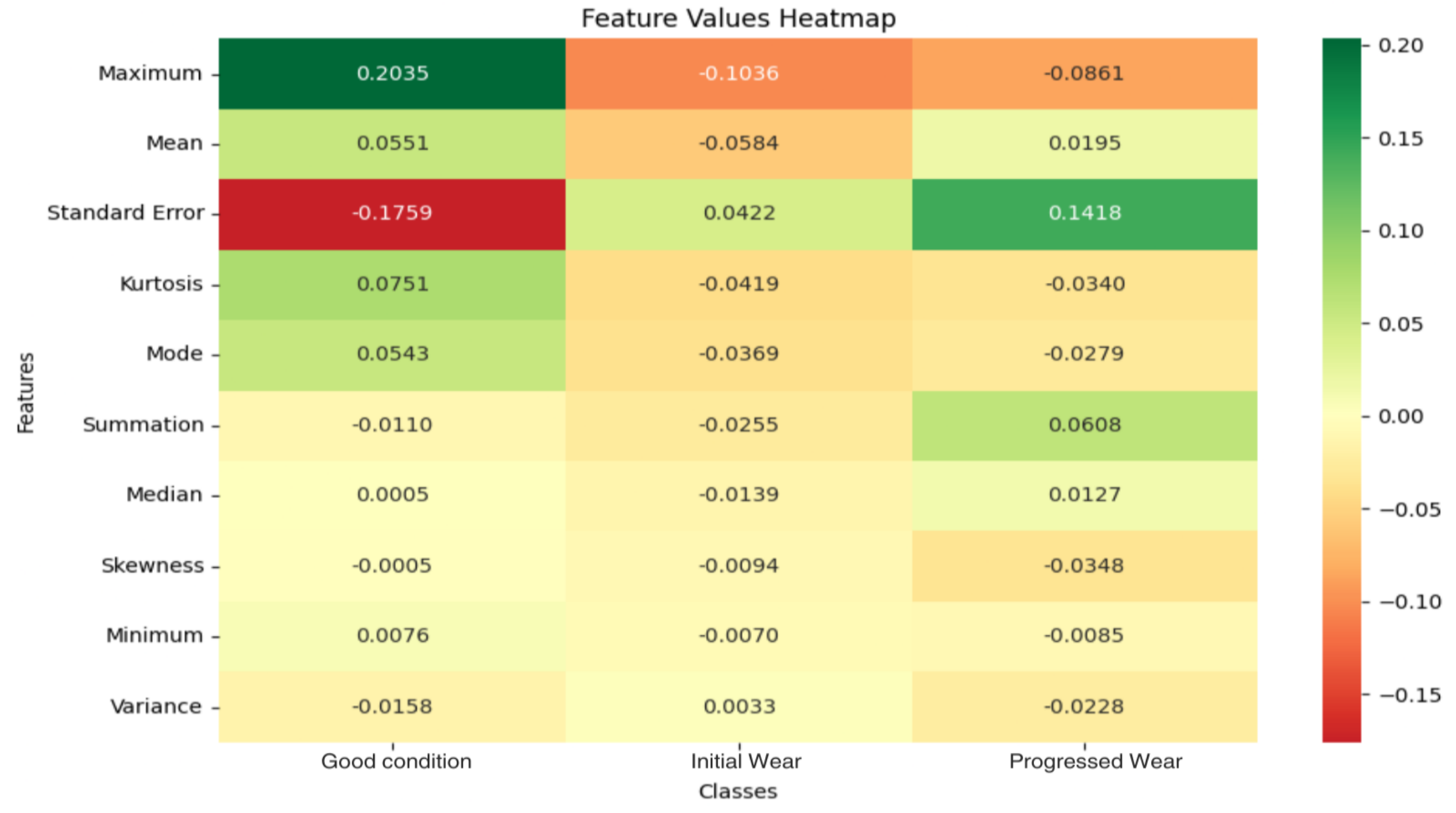

Figure 17. Local representation of Tuned KNN model.

For vanilla KNN

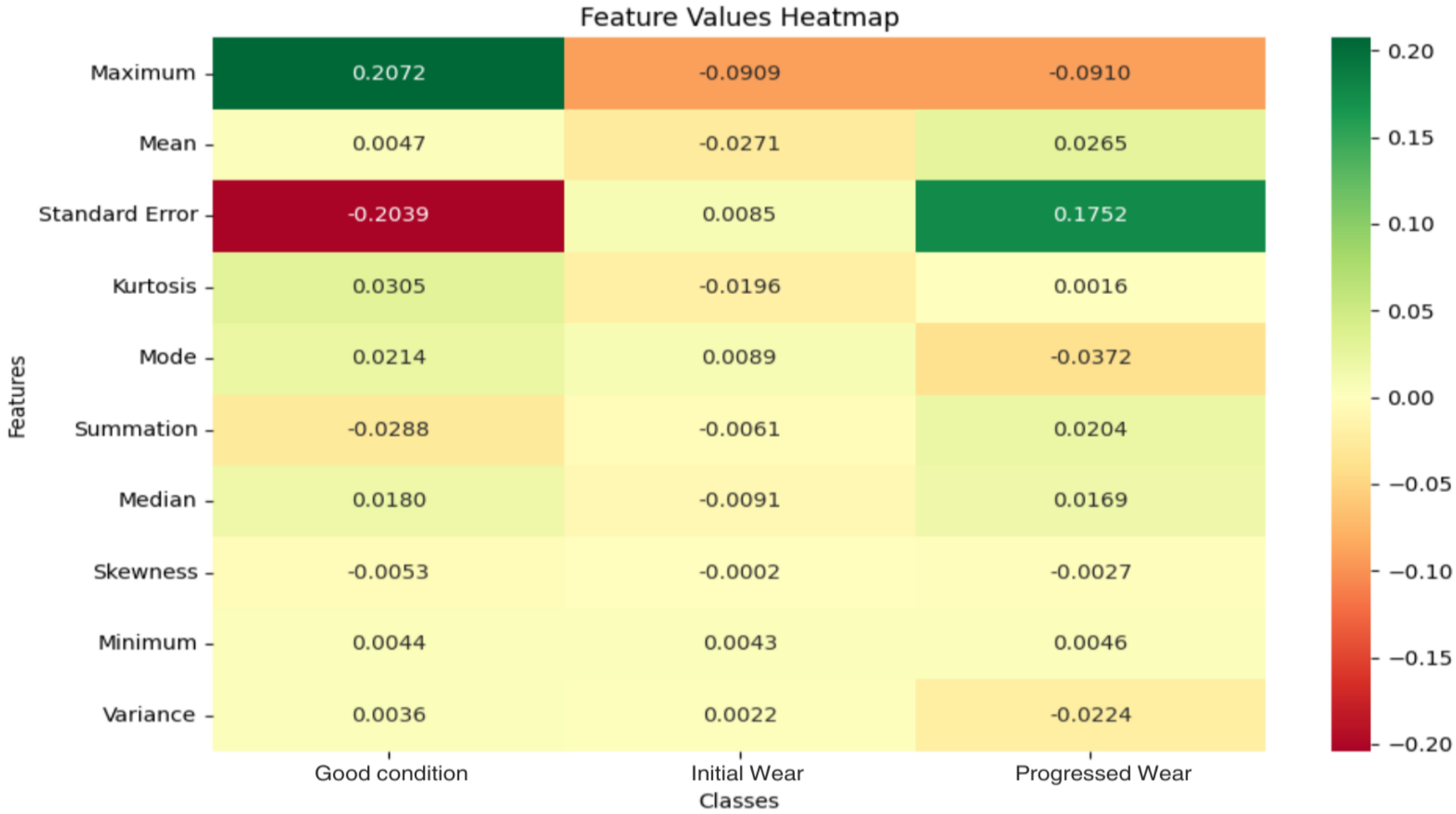

Figure 18. Local representation of Vanilla KNN model.

### 7.2.3. Model performance on unseen data

Fig. 19 depicts a Graphical User Interface (GUI) of the TCMs. It is used to demonstrate the model's performance on unseen data. After entering the values of features, the model is run at the backend, and the corresponding condition of the tool is displayed to the user. This approach is implemented to increase the simplicity with which Tool Condition Monitoring can be performed.

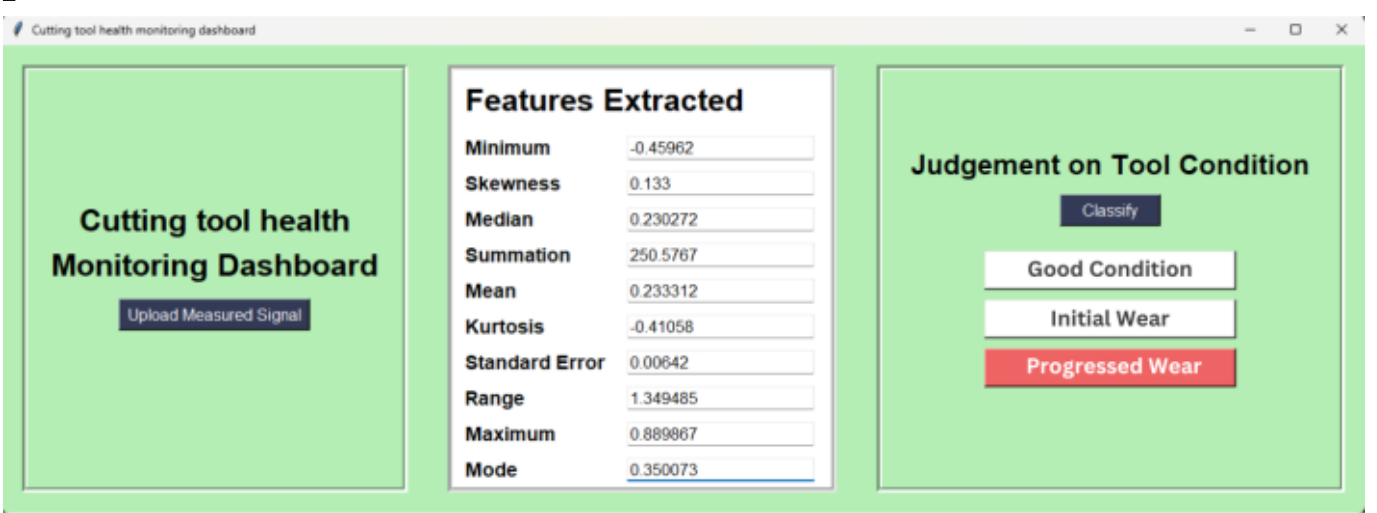

Figure 19. Graphical User interface of the TCMs.

### 7.3. Root cause analysis of type II errors

In the context of Tool Condition Monitoring, Type II errors (false negatives), where a worn or faulty tool is misclassified as being in acceptable condition, are particularly critical as they may lead to tool failure, poor surface quality, and unplanned downtime. The occurrence of Type II errors in this study can be attributed to several underlying factors related to the characteristics of milling force signals and feature space overlap. One primary cause of Type II errors is the overlapping statistical characteristics of force signals between adjacent tool wear states, particularly between the Initial Wear and Bad classes. During the early stages of tool degradation, changes in cutting forces are often subtle and may remain within the variability range of a healthy tool, making clear separation challenging. This overlap is further intensified by the intermittent nature of milling, where variations in chip load and tool–workpiece engagement introduce inherent signal fluctuations. Another contributing factor is local neighborhood ambiguity in the feature space, which is inherent to instance-based classifiers such as KNN. When a test instance lies near the boundary between two wear classes, its nearest neighbors may belong to a less severe wear state, leading to misclassification and increased Type II error. The proposed methodology addresses these root causes through data augmentation and hyperparameter tuning, which enhance class representation and neighborhood density in critical regions of the feature space. Augmentation improves the model's exposure to wear-transition patterns, while distance-weighted KNN emphasizes closer neighbors that better reflect the true tool condition. As demonstrated in the results, these measures significantly reduced Type II error, particularly for force signals in the X-direction. This analysis indicates that Type II errors in milling TCM are primarily driven by signal overlap and transitional wear behavior rather than model inadequacy. By strengthening data representation and local decision reliability, the proposed framework effectively mitigates these errors while maintaining interpretability.

### 7.4. Manufacturing and maintenance implications of the white-box framework

The proposed model-agnostic white-box framework provides both global feature importance and local instance-level explanations that can support manufacturing decision-making and maintenance planning. The statistical force features highlighted by the analysis reflect changes in cutting behavior and may be interpreted in relation to tool condition and machining stability. For example, an increase in skewness of cutting force signals indicates greater asymmetry in the force distribution, which may be associated with uneven material removal, localized edge wear, or increased friction at the tool–workpiece interface. As tool wear develops, force signals can exhibit more pronounced deviations in one direction, leading to higher skewness values. From a manufacturing perspective, consistently elevated skewness may therefore suggest changes in cutting conditions that warrant closer observation or inspection of the cutting tool. Similarly, variations in the mode of force-related features provide insight into process stability. The mode represents the most frequently occurring force value during machining. A decrease in the mode, particularly when accompanied by increased variance or range, suggests that the cutting process spends less time operating under relatively stable force conditions and more time experiencing fluctuating loads. Such behavior is commonly linked to tool wear, early stages of chatter, or changes in tool–workpiece contact conditions. In practice, these trends may indicate the need for corrective actions to maintain machining stability. Other features identified through the white-box analysis, such as variance, range, and kurtosis, further contribute to the interpretation of tool condition. Increased variance and range reflect greater force fluctuations that may arise from worn cutting edges and increased friction, while elevated kurtosis indicates the presence of occasional force spikes associated with intermittent cutting or minor edge damage. When considered collectively, these features provide a broader and more interpretable view of tool condition than single metrics alone. The local explanations generated by LIME support instance-level analysis by indicating which features contribute most to a specific classification outcome. For example, if a classification indicating a degraded tool condition is primarily influenced by skewness and kurtosis, tool inspection or replacement may be considered. Conversely, if a prediction is driven mainly by features related to force magnitude, adjustments to machining parameters may be evaluated before changing the tool. This distinction can help reduce unnecessary tool replacements and support more informed maintenance decisions. From a maintenance planning perspective, monitoring trends in interpretable features across machining cycles can support condition-based maintenance strategies. Rather than relying solely on fixed tool life intervals, maintenance actions may be initiated when multiple wear-sensitive features show consistent deviations from normal operating behavior. In terms of tool change thresholds, white-box outputs allow thresholds to be defined based on the combined behavior of several physically meaningful features, which may improve robustness compared to single-feature criteria. For predictive maintenance, gradual changes in dominant features, together with increasing influence of wear-related indicators in local explanations, can signal an approaching end-of-life condition. This information may be used to schedule tool replacement in advance, reducing the likelihood of unexpected tool failure and supporting stable production planning. Overall, by providing interpretable insights into model decisions, the white-box framework helps connect data-driven predictions with practical manufacturing and maintenance considerations.

## 8. Conclusions

This study presented a KNN-based Tool Condition Monitoring (TCM) framework using real-time milling force signals combined with a model-agnostic white-box interpretability approach. Statistical feature extraction and feature selection were applied to force signals, followed by classification using both vanilla and tuned KNN models. A comparative analysis of force signals revealed that the augmented force data in the X-direction provided superior classification performance compared to the Y-direction, making it more suitable for reliable tool condition assessment. Type II errors, which are critical in TCM applications, were significantly reduced through data augmentation and distance-weighted KNN, improving decision reliability. The impact of hyperparameter tuning was demonstrated by comparing vanilla and tuned KNN models, with the tuned model achieving improved generalization and reduced misclassification. Furthermore, the proposed model-agnostic white-box framework enhanced interpretability by providing both global insights into dominant force features and local, instance-level explanations of individual predictions, supporting informed tool maintenance decisions. Tool condition la-beling in this study was performed based on controlled machining progression and visual in-spection of the cutting tool, following commonly adopted practices in milling TCM. Although international standards such as ISO 8688 provide guidelines for quantitative tool life evaluation, direct flank wear measurements were not continuously available during force signal acquisition. Experimental repeatability verification through repeated trials under identical wear states was not conducted and is acknowledged as a limitation. Future work will focus on standardized wear calibration and repeatability analysis to further strengthen the robustness and generalizability of the proposed framework. Overall, the integration of KNN classification with a model-agnostic interpretability layer demonstrates a practical and transparent approach for advancing force-based Tool Condition Monitoring in milling operations.

**Nomenclature and Abbreviations**

| **Abbreviations** | |
|---|---|
| TCM | Tool Condition Monitoring |
| ML | Machine Learning |
| KNN | K-Nearest Neighbors |
| CNC | Computer Numerical Control |

| HSS | High-Speed Steel |
|---|---|
| SVM | Support Vector Machine |
| CNN | Convolutional Neural Network |
| PCA | Principal Component Analysis |
| ANFIS | Adaptive Neuro-Fuzzy Inference System |
| RF | Random Forest |
| FURIA | Fuzzy Unordered Rule Induction Algorithm |
| GUI | Graphical User Interface |
| LIME | Local Interpretable Model-Agnostic Explanations |
| DAQ | Data Acquisition |
| rpm | Revolutions Per Minute |
| Fx | Cutting Force in X-direction |
| Fy | Cutting Force in Y-direction |
| Fz | Cutting Force in Z-direction |
| ISO | International Organization for Standardization |
| ROC AUC | Receiver Operating Characteristic – Area Under Curve |
| TP | True Positive |
| FP | False Positive |
| DMLP | Deep Multi-Layer Perceptron |
| NN | Neural Network |
| CV | Cross-Validation |

| **Symbols** | |
|---|---|
| $x$ | Input feature vector |
| $x_i$ | i-th feature in the input vector |
| $\mathbf{x}$ | Multivariate feature vector |
| $\mathbf{x}'$ | Augmented feature vector |
| $y$ | True tool condition label |
| $\hat{y}$ | Predicted tool condition label |
| $f(\cdot)$ | Trained prediction model (black-box model) |
| $g(\cdot)$ | Interpretable surrogate model (LIME) |
| $n$ | Number of features |
| $k$ | Number of nearest neighbors in KNN |
| $d(\cdot,\cdot)$ | Distance function |
| $d_i$ | Distance to the i-th nearest neighbor |
| $w_i$ | Weight of the i-th neighbor |
| $\epsilon$ | Feature-level perturbation vector |
| $z$ | Perturbed sample used in LIME |
| $w_i$ | Feature weight in local surrogate model |
| $L(f, g, \pi_{x_0})$ | Fidelity loss between black-box and surrogate model |
| $\pi_{x_0}$ | Proximity measure around instance $x_0$ |
| $\Omega(g)$ | Complexity penalty enforcing interpretability |
| $x_0$ | Instance being explained by LIME |
| $Fx$, $Fy$, $Fz$ | Cutting force in feed (X, Y, Z) directions |
| $d(P_1, P_2)$ | Manhattan distance between two points |
| $(x_1, x_2), (y_1, y_2)$ | Coordinates of point $P_1$ and $P_2$ |
| $\mu$ | Mean value of force signal |
| $\sigma$ | Standard deviation of force signal |